\documentclass{article}
\usepackage{spconf,amsmath,graphicx}
\usepackage{array,multirow}
\usepackage{cite}
\usepackage{enumerate}
\usepackage{paralist}

\title{Integrating Text Inputs For Training and Adapting \\ RNN Transducer ASR Models}
\name{\begin{tabular}{c} Samuel Thomas, Brian Kingsbury, George Saon, Hong-Kwang J. Kuo \end{tabular}}
\address{\begin{tabular}{c} IBM Research AI \end{tabular}}

\begin{document}
\ninept
\maketitle
\begin{abstract}
Compared to hybrid automatic speech recognition (ASR) systems that use a modular architecture in which each component can be independently adapted to a new domain, recent end-to-end (E2E) ASR system are harder to customize due to their all-neural monolithic construction. In this paper, we propose a novel text representation and training framework for E2E ASR models. With this approach, we show that a trained RNN Transducer (RNN-T) model's internal LM component can be effectively adapted with text-only data. An RNN-T model trained using both speech and text inputs improves over a baseline model trained on just speech with close to 13\% word error rate (WER) reduction on the Switchboard and CallHome test sets of the NIST Hub5 2000 evaluation. The usefulness  of the proposed approach is further demonstrated by customizing this general purpose RNN-T model to three separate datasets. We observe 20-45\% relative word error rate (WER) reduction in these settings with this novel LM style customization technique using only unpaired text data from the new domains.
\end{abstract}

\section{Introduction}

With their remarkable performance and simplified training pipeline, end-to-end (E2E) models have become the de-facto approach to automatic speech recognition, replacing traditional hybrid ASR models that separately model different knowledge sources such as the language model, acoustic model, and lexicon. One advantage to the modular architecture of hybrid models, however, is that each component can be trained or adapted on separate, independent data sets. On the other hand, current all-neural E2E systems require transcribed training sets with paired speech and text transcripts. This limitation becomes profound, especially when ASR models need to be customized for new domains. With hybrid models, domain adaptation could be performed by adapting the language model on task- or domain-specific text-only data and updating the lexicon to cover any new words. We address this shortcoming of E2E models using a novel representation that effectively integrates text inputs into model training and adaptation.

\begin{figure}
    \centering
    \includegraphics[scale=0.4]{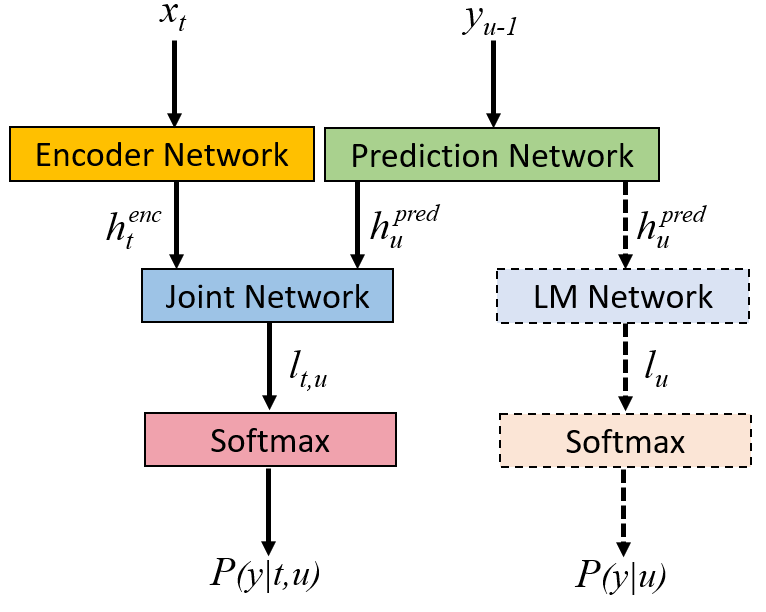}
    \caption{Architecture of an RNN-T transducer ASR model with additional layers for NN-LM customization}
    \label{fig:arch}
\end{figure}

To demonstrate the usefulness of our method we conduct a study on RNN-Transducer (RNN-T) models, a class of end-to-end, streamable, all-neural models that have been widely adopted for speech recognition~\cite{he2019streaming,rao2017exploring,li2019improving,shafey2019joint,ghodsi2020rnn}. RNN-T models typically consist of three different sub-networks: an encoder network, a prediction network, and a joint network~\cite{graves2012sequence} as shown in Figure~\ref{fig:arch}. The encoder network  or transcription network produces acoustic embeddings, while the prediction network resembles a language model in that it is conditioned on previous non-blank symbols produced by the model. The joint network combines the two embedding outputs to produce a posterior distribution over the output symbols. This architecture elegantly replaces a conventional hybrid ASR system composed of separate acoustic model, language model, pronunciation lexicon, and decoder components.

Several methods have been proposed to adapt or customize these models to new domains. An external language model can be trained separately on data specific to the new task and then integrated with the existing RNN-T model via shallow fusion, density ratio fusion~\cite{mcdermott2019densityratio}, or internal language model estimation~\cite{meng2021ilme}. Because scores from the external language model are combined with scores from the RNN-T model during decoding, a more complex decoder is necessary. Other methods attempt to directly update one or more parts of the RNN-T model, and thus do not require changes to decoding. A text-to-speech (TTS) model can be used to synthesize audio from text, and then the paired synthetic speech and text can be used to adapt the prediction network \cite{kurata2021improving}. A more recent approach avoids synthesizing audio from text. This method first inserts a temporary LM layer to the prediction network as shown in Figure~\ref{fig:arch}. With this new layer, the prediction network is then trained as a neural LM on external text data \cite{pylkkonen2021fast}.

\section{Textograms}

\begin{figure}
    \centering
    \includegraphics[scale=0.45]{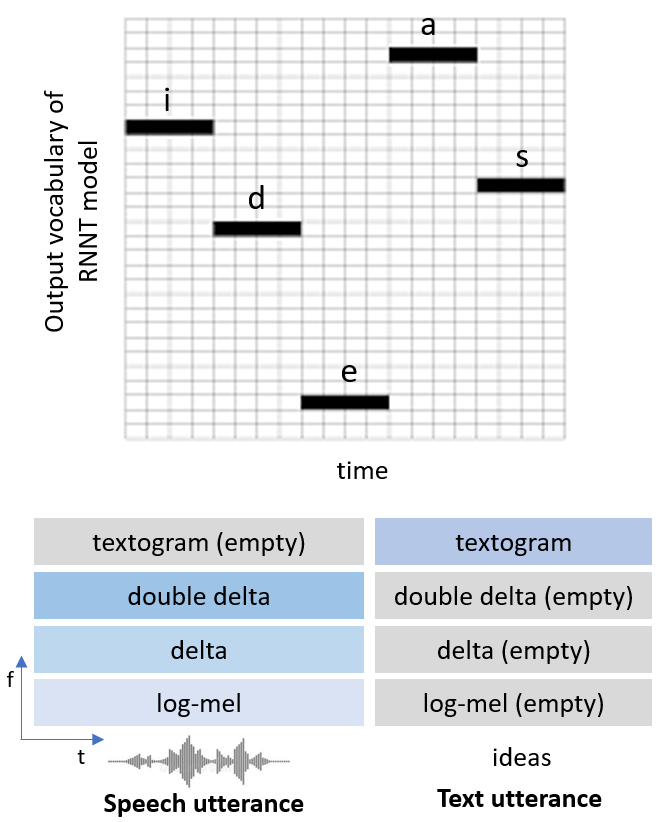}
    \caption{Schematics of the textogram representation for an input text input, `\textit{ideas}' (top), feature inputs used to train RNN-T models corresponding to speech and text (bottom).}
    \label{fig:textogram}
\end{figure}

In this work, we propose a novel approach to adapting RNN-T models to domains and tasks unseen during training. Like~\cite{pylkkonen2021fast}, our method uses only text data, does not require TTS, and does not require a more complex decoder that combines scores from multiple models. Unlike~\cite{pylkkonen2021fast}, our approach optimizes the RNN-T loss during adaptation. The main idea we use is to train the transcription network to work with two different input modalities: the standard acoustic features (log-Mel spectrograms augmented with $\Delta$ and $\Delta^{2}$ features in this work) and also a representation we call a ``textogram''. A textogram is simply a concatenation of the one-hot encodings of the symbols making up the reference text, with the encoding of each symbol being repeated some number of times such that the textograms have durations similar to spectrograms. For example, given a reference text \textit{ideas} and an RNN-T model that operates on graphemes, textogram features are constructed by first splitting the word into its constituent graphemes, \textit{i}, \textit{d}, \textit{e}, \textit{a}, and \textit{s}. Each symbol is assigned a duration, four frames in this case, to create a 2-dimensional representation as shown in Figure~\ref{fig:textogram}. This representation is very similar to posteriograms, which are symbol posterior probability estimates from a trained neural network acoustic model. 

Once constructed in this fashion, these representations are used along with traditional speech based log-mel features to train RNN-T models. In this work, we simply increase the dimensionality of the input to the first layer of the transcription network to be the sum of the dimensionality of the acoustic features and the textogram features, and then fix the unused modality's inputs to $0.0$ during training, but other designs are certainly possible. 

To allow the model to learn robustly from the textogram representations, various variabilities can be added to this representation. These choices include:
\begin{enumerate}[(a)]
\item Label masking: To allow the model to learn useful n-gram sequences instead of blindly memorizing the text, active entries of the textogram representation can be randomly dropped. The rate of label masking is a parameter that can be empirically selected.
\item Label confusions: The acoustic confusion among various speech sounds can be introduced into the textogram by substituting various labels with their confusable sounds e.g., \textit{p} and \textit{b} 
\item Variable label duration: The length of each linguistic unit can be varied to model real durations in the speech signal. In Figure~\ref{fig:textogram} we use four frames per symbol.
\item Modeling pronunciations: The input textogram may include different ``sounds-like'' sequences for a given target output. For example, the target \textit{Miami}, may be associated with textogram sequences, \textit{Miami}, \textit{my Amy}, or \textit{mee Amy}. 
\item Multiple linguistic symbol sets:  The symbol set used with textograms can be different from the output symbol set for the transducer model. For example, phonetic targets can be used at the RNNT's output while graphemes are used for textograms.
\end{enumerate}
In this work, we use a fixed label duration for various text symbols along with label masking, to construct textogram features

\section{Training and Adapting RNN-T Models}
\subsection{Training RNN-T models}
Using notation from \cite{graves2012sequence}, an RNN-T models the conditional distribution $p(\mathbf{y}|\mathbf{x})$ of an output sequence $\mathbf{y}=(y_1,\dots,y_U) \in \mathcal{Y}^*$ of length $\mathcal{U}$ given an input sequence $\mathbf{x}=(x_1,\dots,x_T) \in \mathcal{X}^*$ of length $T$. In an ASR setting, while the elements of $\mathbf{x}$ are continuous multidimensional speech features, $\mathbf{y}$ is discrete and corresponds to an output symbol set, like the grapheme set of the language being modelled by the network. To facilitate the alignment of the two sequences, which in general have different lengths, the output alphabet is augmented with an additional BLANK symbol. 
$p(\mathbf{y}|\mathbf{x})$ is computed by marginalizing over all possible alignments between $\mathbf{x}$ and $\mathbf{y}$.

The probability of a particular alignment is computed in terms of embeddings, $h^{enc}$, of the input sequence computed by the encoder network and embeddings, $h^{pred}$, of the output sequence computed by a prediction network.  The joint network combines these two embeddings to produce a posterior distribution over the output symbols. Training is based on an efficient forward-backward algorithm, with $T \times U$ complexity for both loss and gradient computation, that minimizes $-\log p(\mathbf{y}|\mathbf{x})$, the negative log-likelihood loss and uses speech data paired with corresponding text transcripts.
Representing text as textograms allows us to extend the RNN-T training framework. In addition to training the network using samples comprising paired speech and text data, $(\mathbf{x}^{sp},\mathbf{y})$, where the speech is represented by an acoustic feature sequence $\mathbf{x}^{sp}$ and the text is represented by a symbol sequence $\mathbf{y}$, we also train the network using samples comprising paired text representations $(\mathbf{x}^{txt}, \mathbf{y})$, where $\mathbf{x}^{txt}$ is the textogram representation of the text and $\mathbf{y}$ is the symbol sequence. 

For the (speech, text) samples in the training set, we extract log-mel features augmented with $\Delta$ and $\Delta^{2}$ features and set the input dimensions corresponding to the textogram features to $0.0$, as shown in Figure~\ref{fig:textogram}. To improve the robustness of the speech training, sequence noise injection~\cite{saon2019seqnoise} and SpecAugment~\cite{park2019specaugment} are applied to the speech features. 
For the (text, text) samples, we compute textogram features for each transcript and set the input dimensions corresponding to the acoustic features to $0.0$, as shown in Figure~\ref{fig:textogram}. To prevent the (text, text) task from being completely trivial, we apply label masking to the textogram features.
By integrating text inputs into the training pipeline, the RNN-T model's transcription network is now trained as a single encoder for two modalities: speech and text. With this joint training, the transcription network produces similar embeddings for both speech and text that can be further used along with a prediction and joint network that are shared by both modalities.

\subsection{Adapting RNN-T models}

Once an RNN-T model has been trained on both speech and text, it can be easily be adapted to a new domain using only text data.  Prior to the adaptation process, the text-only adaptation data is converted into textogram features. The RNN-T model is then adapted using these features. The parameters of the transcription network are kept constant during the adaptation process, while the parameters of the prediction and joint networks may be updated. This ensures that the model is still acoustically robust while  being able to effectively process data from the new domain.

\section{Experiments and Results}

\subsection{The RNN-T base model with textogram features}

The RNN-T models used in our experiments are trained on a collection of US English telephony data including Switchboard, Fisher, and proprietary data. Each RNN-T model has several sub-networks as illustrated in Figure~\ref{fig:arch}. The transcription network  contains 6 bidirectional LSTM layers with 640 cells per layer per direction. The prediction network is a single unidirectional LSTM layer with only 1024 cells. The joint network projects the 1280-dimensional stacked encoder vectors from the last layer of the transcription net and the 1024-dimensional prediction net embedding each to 256 dimensions, combines them multiplicatively, and applies a hyperbolic tangent. Finally, the output is projected to 42 logits, corresponding to 41 characters plus BLANK, followed by a softmax. More details on training settings and design choices can be found in \cite{george2021rnn}. The RNN-T based ASR models are trained using 40-dimensional,  global mean and variance normalized log-Mel filterbank features, extracted every 10 ms. These features are augmented with  $\Delta$ and $\Delta^{2}$ coefficients, every two consecutive frames are stacked, and every second frame is skipped, resulting in 240-dimensional vectors every 20 ms. Likewise, every two frames of the textogram representations are stacked, and every second frame is skipped, resulting in 84-dimensional vectors every 20 ms. Thus, the transcription network takes a 324-dimensional input. The speech data is augmented using speed and tempo perturbation with values in \{0.9, 1.1\} for both speed and tempo separately, resulting in 4 additional speech training data replicas. For sequence noise injection, we add, with probability 0.8, to the spectrum of each training utterance the spectrum of one random utterance of similar length scaled by a factor of 0.4. For SpecAugment we used the settings published in \cite{park2019specaugment}.

Textogram representations of the text data are generated using the same grapheme set that is modelled at the outputs of the RNN-T and a fixed duration of four frames per symbol. Label masking at a rate of 25\% is applied to the textograms to prevent the model from simply reproducing the input. 

The RNN-T models are  trained in Pytorch on V100 GPUs for 20 epochs using an AdamW optimizer. The maximum learning rate is set to 2e-4 and the OneCycleLR policy consists in a linear warmup phase from 2e-5 to 2e-4 over the first 6 epochs followed by a linear annealing phase to 0 for the remaining 14 epochs. We use an effective batch size of 128 utterances. Batches are constructed from feature sequences of similar lengths without regard to whether the features are mel spectrograms or textograms, so generally each batch will contain both types of training sample.

We train two RNN-T models in our first set of experiments: an RNN-T model on all the available speech data and a textogram based model trained on both speech and text. Both the models have the same architecture, except that the first layer in the transcription network has a larger input in the model trained on speech and text, and both models are constructed using the same training procedure  described above. In Table \ref{tab:table_1} we report results on the commonly
used Hub5 2000 Switchboard (SWB) and CallHome (CH) test sets, which are processed using LDC segmentations and scored using Kaldi scoring setups for measuring WER. An RNN-T model trained with the proposed textogram (TOG) method, TOG-RNN-T, significantly improves over a competitive baseline model trained on just speech data. We hypothesize that the relative WER reduction of 10\% and 13\% on SWB and CH using the model jointly trained with speech and text inputs is due to a regularization effect caused by training on two modalities using twice as much data. The twofold increase in training data is because we use the transcripts corresponding to the speech data as additional text inputs to train the model. From one perspective, then, using textograms can be seen as another approach to data augmentation when training RNN-T models for ASR.
\begin{table}[tbph]
    \centering
    \begin{tabular}{|c|c|c|}
    \hline
    \textbf{Model} & \textbf{SWB} & \textbf{CH}\\
    \hline
    RNN-T & 6.9 & 11.9 \\
    \hline
    TOG-RNN-T & 6.2 & 10.5 \\
    \hline
    \end{tabular}
    \caption{Baseline WER on Hub5’00 Switchboard and CallHome using only speech data (RNN-T) and the proposed approach with both speech and text inputs (TOG-RNN-T).}
    \label{tab:table_1}
\end{table}
\vspace{-0.7cm}
\subsection{RNN-T adaptation to various domains}

In our next set of experiments, we adapt the general purpose textogram based model to various new domains and settings. To measure the usefulness of our proposed technique we also implement the text-only adaptation technique proposed in \cite{pylkkonen2021fast} and compare results. 

In \cite{pylkkonen2021fast}, the prediction network is interpreted as a neural LM. As shown in Figure~\ref{fig:arch}, to adapt the prediction network on text-only input, a temporary LM layer is first attached and trained along with a softmax output layer on training data transcripts using a standard  cross-entropy  loss. Once the LM layer has been trained, it is kept fixed while the prediction network is further adapted to novel text data from the new domain. This training uses  two auxiliary losses for better regularization: a KL divergence loss that controls how similar the adapted model's output distribution is to the original unadapted model's distribution and a weight regularization loss that prevents the adapted model's weights from drifting away from the base model. Compared to this NN-LM based technique, our proposed textogram based method performs adaptation by optimizing the RNN-T loss rather than cross-entropy. We conduct our adaptation experiments on three diverse datasets. \newline
\textbf{A. Adaptation to the SLURP dataset}: The SLURP dataset \cite{bastianelli2020slurp} is a recently released dataset for spoken language understanding. The audio data was collected in challenging acoustic conditions found in a typical home or office environment using far-field and close speaking microphones. The training partition of the dataset has about 11K text sentences. We adapt the base TOG-RNN-T model using text-only data from this dataset and test it on the corpus's speech test set, which corresponds to 10 hours of speech. The test data is also downsampled to 8~kHz for our use. Given that the SLURP dataset was collected for developing an in-home personal robot assistant, this domain is quite different from the original base model training data. 
Table 2 shows the performance of various adaptation techniques on this dataset.  The unadapted model's WER is quite high since the SLURP dataset is substantially different both acoustically and lingustically. The neural LM based adaptation technique \cite{pylkkonen2021fast} reduces the WER by about 15\% relative. We next adapt the RNN-T model in three different ways using our proposed method:
\begin{inparaenum}[(1)]
\item in \textit{TOG adapt (P+J)} we adapt both the prediction and joint networks,
\item in \textit{TOG adapt (P)} we adapt only the prediction network, and
\item in \textit{TOG adapt (P) + NN-LM} we combine textogram adaptation with the NN-LM technique.
\end{inparaenum}
Adapting the RNN-T base model with textogram based features significantly improves the WER reduction. We observe more gains by adapting just the prediction network than adapting both the prediction and the joint networks. Combining the NN-LM adaptation method provides further regularization to the textogram based adaptation and provides the best relative WER reduction of 23\%.

\begin{table}[htbp]
    \centering
    \begin{tabular}{|l|c|}
    \hline
    \textbf{Model} & \textbf{WER\%}\\
    \hline
    Unadapted TOG-RNN-T & 47.5\\
    \hline
    NN-LM adapt \cite{pylkkonen2021fast}  & 40.4 \\
    \hline
    TOG adapt (P+J)  & 38.6 \\
    \hline
    TOG adapt (P) & 37.6 \\
    \hline
   TOG adapt (P) + NN-LM & 36.5 \\
    \hline
    \end{tabular}
    \caption{Performance of various models adapted to SLURP.}
    \label{tab:table_2}
\end{table}

\noindent \textbf{B. Adaptation to the ATIS dataset}: In our second set of experiments we use the ATIS \cite{hemphill1990atis} training and test sets: 4976 training utterances from Class A (context independent) training data in the ATIS-2 and ATIS-3 corpora and 893 test utterances from the ATIS-3 Nov93 and Dec94 data sets.  The test utterances comprise about 1.5 hours of audio from 55 speakers. The data was originally collected at 16~kHz, but is downsampled to 8~kHz to match the telephony base model. We repeat the same set of adaptation experiments on this dataset as well. Although the WER of the unadapted model is much lower compared to results on SLURP, the model still benefits from adaptation. Similar to previous results, the NN-LM adaptation technique improves WER reduction by 30\% relative. Although the proposed textogram adaptation of just the prediction network improves on top of these gains to 40\% relative WER reduction, combining with the NN-LM  improves the performance further by 45\% relative.

\begin{table}[htbp]
    \centering
    \begin{tabular}{|l|c|}
    \hline
    \textbf{Model} & \textbf{WER\%}\\
    \hline
    Unadapted TOG-RNN-T & 3.1\\
    \hline
    NN-LM adapt \cite{pylkkonen2021fast} & 2.2 \\
    \hline
    TOG adapt (P+J) & 1.8 \\
    \hline
    TOG adapt (P) & 1.9  \\
    \hline
   TOG adapt (P) + NN-LM & 1.7  \\
    \hline
    \end{tabular}
    \caption{Performance of various models adapted to ATIS.}
    \label{tab:table_3}
\end{table}

\noindent \textbf{C. Adaptation to the Harper Valley dataset}: In our final set of experiments, we adapt the baseline RNN-T model to the Harper Valley Bank corpus \cite{wu2020harpervalleybank}. This dataset is a public domain corpus with spoken dialogs that simulate simple consumer banking interactions between users and agents. There are 1,446 human-human conversations between 59 unique speakers. We adapt the model on 15K text transcripts from the training partition used in \cite{ganhotra2021integrating} before testing the adapted model on a 1.5 hour speech test set.

\begin{table}[htbp]
    \centering
    \begin{tabular}{|l|c|}
    \hline
    \textbf{Model} & \textbf{WER\%}\\
    \hline
    Unadapted TOG-RNN-T & 10.4\\
    \hline
    NN-LM adapt \cite{pylkkonen2021fast} & 7.5 \\
    \hline
    TOG adapt (P+J) & 7.3 \\
    \hline
    TOG adapt (P) & 7.0 \\
    \hline
   TOG adapt (P) + NN-LM & 6.8 \\
    \hline
    \end{tabular}
    \caption{Performance of various models adapted to Harper Valley.}
    \label{tab:table_4}
\end{table}

Similar to the previous two customization experiments, we see significant  performance gains using the proposed adaptation techniques. Consistent gains are observed by adapting just the prediction network using the text-only transcripts. Compared to the NN-LM method, the textogram based method is able to better adapt this sub-network. We hypothesize that this is because of the holistic nature of the adaptation process where the prediction network is not adapted independently using a separate loss, but is adapted along with embeddings from the transcription network using the original RNN-T training loss, via the joint network. In this case as well, we see up to 34\% relative WER reduction after adaptation compared to the unadapted model. Combining the NN-LM training loss with the RNN-T training loss gives the best improvements, suggesting that the prediction network can benefit further when good regularization is used. 

For all our adaptation experiments, we train for 20 epochs using an AdamW optimizer and a OneCycleLR policy set to a maximum learning rate of 2e-4. 
In our experiments where we integrate the NN-LM training loss with the RNN-T training loss, we weight the NN-LM loss by an empirically determined weight of 2e2 before combining it with the original RNN-T loss. 
While generating textogram features in both train and adaptation, certain parameters have been selected intuitively. The frame duration of each constituent symbol in the textogram representation is currently set to 4 as a close approximation to an average speaking rate of 4-5 syllables (12-15 symbols) per second. Similarly, while stacking feature frames, we stack textogram features in exactly the same way that we stack speech features. We have also not yet explored the use of label confusions, multiple linguistic symbol sets or modeling pronunciations. These parameters and settings need be explored further.

In our final set of experiments we compare the performance of the adapted TOG-RNN-T model with our baseline RNN-T model combined with an external LM. The external NN-LM is a two layer LSTM with 1024 nodes trained on the same adaptation data used to adapt the TOG-RNNT model. The scores from the external LM and the RNN-T model are combined via shallow fusion \cite{george2021rnn}. Although improvements with shallow fusion are observed (row 2 vs. row 1 of Table \ref{tab:table_5}), the gains using the TOG model are equally significant (row 3).  Compared to the shallow fusion approach, our proposed method has several benefits. Adapting the prediction network via textograms does not require a more complex decoder that combines scores from multiple models, while at the same time providing better performance. The results with shallow fusion were obtained after carefully tuning the external LM weight.  The need to optimize the LM weight is also eliminated with our method.

\begin{table}[htbp]
    \centering
    \begin{tabular}{|l|c|c|c|}
    \hline
    \textbf{Model} & \textbf{SLP} & \textbf{ATIS} & \textbf{HVB}\\
     \hline
    Unadapted RNN-T & 51.3 & 3.6 & 11.3\\
    \hline
    + Shallow Fusion & 41.8 & 1.9 & 7.0 \\
    \hline
   TOG adapt (P) + NN-LM & 36.5 & 1.7 & 6.8 \\
    \hline
    \end{tabular}
    \caption{Comparison of baseline RNN-T WER performance with shallow fusion on SLURP (SLP), ATIS and Harper Valley (HVB).}
    \label{tab:table_5}
\end{table}

\section{Conclusions}
In this paper, we have presented a novel representation for training E2E ASR models with additional text inputs. 
We have shown that this framework reduces the need for paired speech-text data and allows for a simplified adaptation pipeline as well. This work is also novel since we now show how a single ASR encoder can be trained within the RNN-T framework for two modalities: speech and text. The usefulness of the proposed approach is demonstrated by customizing an RNN-T model to three separate datasets with text-only data. We observe 20-45\% relative word error rate (WER) reduction in these settings with this novel technique.

\vfill\pagebreak
\bibliographystyle{IEEEbib}
\bibliography{refs}

\end{document}